\documentclass[sigconf]{acmart}

\usepackage{subfigure}
\usepackage{multirow}
\usepackage{multicol}
\usepackage{diagbox}
\usepackage{bbding}
\usepackage{balance}
\usepackage{setspace}

\AtBeginDocument{%
  \providecommand\BibTeX{{%
    \normalfont B\kern-0.5em{\scshape i\kern-0.25em b}\kern-0.8em\TeX}}}

\copyrightyear{2021} 
\acmYear{2021}
\setcopyright{acmcopyright}
\acmConference[MM '21]{Proceedings of the 29th ACM International Conference on Multimedia}{October   20--24, 2021}{Virtual Event, China}
\acmBooktitle{Proceedings of the 29th ACM International Conference on Multimedia (MM '21), October   20--24, 2021, Virtual Event, China}
\acmPrice{15.00}
\acmDOI{10.1145/3474085.3475238}
\acmISBN{978-1-4503-8651-7/21/10}
\begin{document}
\fancyhead{}
\title{PIMNet: A Parallel, Iterative and Mimicking Network \\ for Scene Text Recognition}

\newcommand\blfootnote[1]{
\begingroup
\renewcommand\thefootnote{}\footnote{#1}
\addtocounter{footnote}{-1}
\endgroup
}
\settopmatter{printacmref=false}

\author{Zhi Qiao$^{1,2}$, Yu Zhou$^{1,2*}$, Jin Wei$^3$, Wei Wang$^{1,2}$, Yuan Zhang$^3$, Ning Jiang$^4$, Hongbin Wang$^4$, Weiping Wang$^1$}
\affiliation{
\institution{$^1$Institute of Information Engineering, Chinese Academy of Sciences, Beijing, China\\$^2$School of Cyber Security, University of Chinese Academy of Sciences, Beijing, China\\$^3$Communication University of China, Beijing, China\\$^4$Mashang Consumer Finance Co., Ltd.}
\city{}
  \country{}}
\email{{qiaozhi, zhouyu, wangwei3456, wangweiping}@iie.ac.cn}
\email{{weijin, yzhang}@cuc.edu.cn, {ning.jiang02, hongbin.wang02}@msxf.com}


\begin{abstract}
  Nowadays, scene text recognition has attracted more and more attention due to its various applications. Most state-of-the-art methods adopt an encoder-decoder framework with attention mechanism, which generates text autoregressively from left to right. Despite the convincing performance, the speed is limited because of the one-by-one decoding strategy. As opposed to autoregressive models, non-autoregressive models predict the results in parallel with a much shorter inference time, but the accuracy falls behind the autoregressive counterpart considerably. In this paper, we propose a Parallel, Iterative and Mimicking Network (PIMNet) to balance accuracy and efficiency. Specifically, PIMNet adopts a parallel attention mechanism to predict the text faster and an iterative generation mechanism to make the predictions more accurate. In each iteration, the context information is fully explored. To improve learning of the hidden layer, we exploit the mimicking learning in the training phase, where an additional autoregressive decoder is adopted and the parallel decoder mimics the autoregressive decoder with fitting outputs of the hidden layer. With the shared backbone between the two decoders, the proposed PIMNet can be trained end-to-end without pre-training. During inference, the branch of the autoregressive decoder is removed for a faster speed. Extensive experiments on public benchmarks demonstrate the effectiveness and efficiency of PIMNet. Our code will be available at https://github.com/Pay20Y/PIMNet.
\end{abstract}





\maketitle

\begin{small}
\begin{spacing}
1
\textbf{\\\\ACM Reference Format:}

\noindent Zhi Qiao, Yu Zhou, Jin Wei, Wei Wang, Yuan Zhang, Ning Jiang, Hongbin Wang and Weiping Wang. 2021. PIMNet: A Parallel, Iterative and Mimicking Network for Scene Text Recognition. In \textit{Proceedings of the 29th ACM International Conference on Multimedia (MM ’21), October 20-24, 2021, Virtual Event, China.} ACM, New York, NY, USA, 10 pages. https://doi.org/10.1145/\\3474085.3475238
\end{spacing}
\end{small}

\blfootnote{* Yu Zhou is the corresponding author.}

\begin{figure}[t]
\begin{center}
\includegraphics[width=0.95\linewidth]{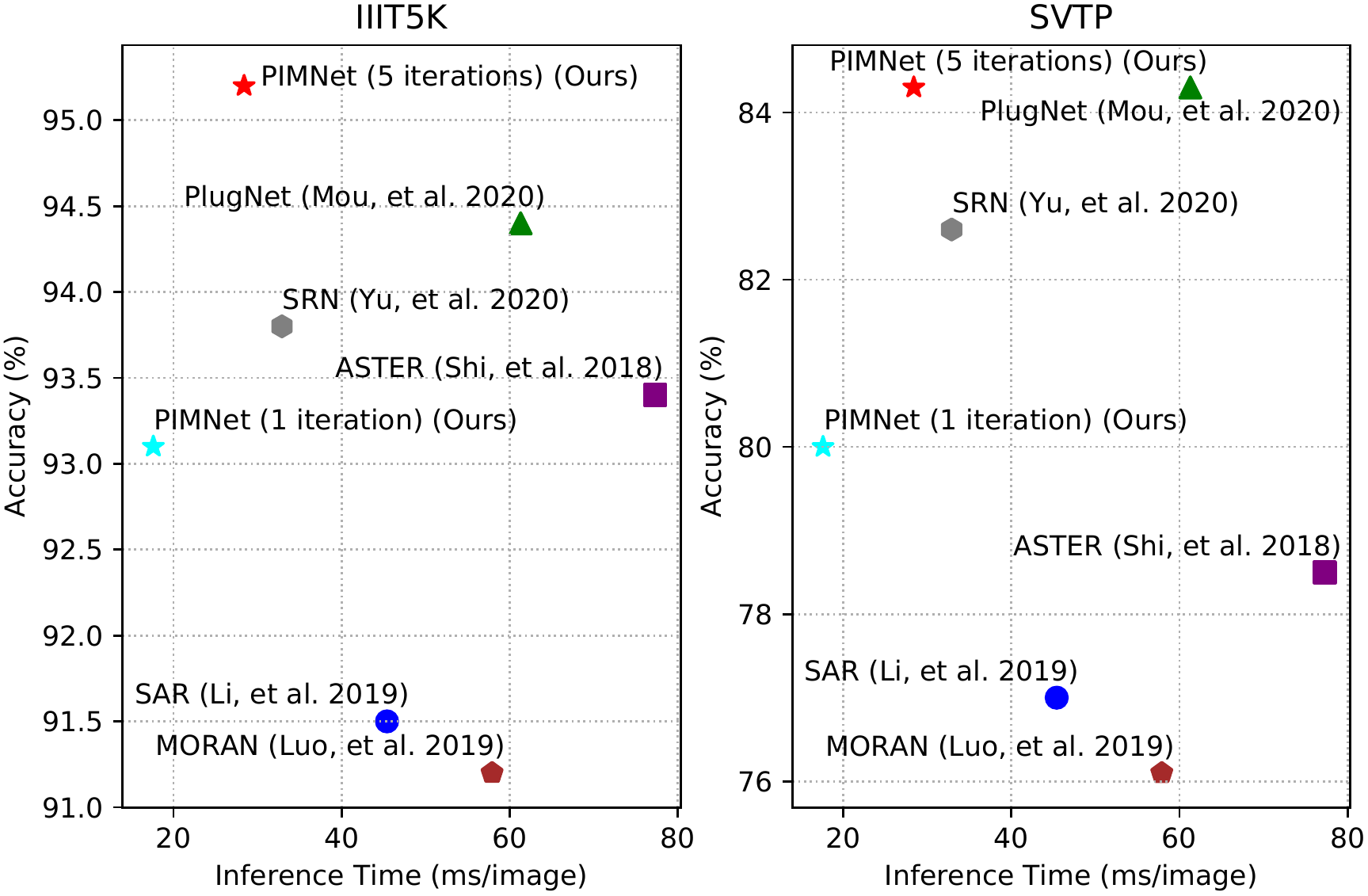}
\end{center}
    \caption{The trade-off illustration between accuracy and inference time (ms/image) on two benchmarks IIIT5K and SVTP of different methods. Our proposed method balances the effectiveness and efficiency. The inference time is measured with the official implementations under the same software and hardware environments.}
    \vspace{-30pt}
\label{fig_acc-time}
\end{figure}

\section{Introduction}
Nowadays, scene text detection and recognition have attracted more and more research interests. Text detection~\cite{liu2014text,tian2016detecting,zhou2017east,qxg1, qxg2, cyd1, cyd2} is the task of localizing the text instances in the scene images. Text recognition is to transcribe the image to editable text format, and scene text recognition is one of the most difficult tasks due to the variety of characters and backgrounds. With the development of deep learning, recent methods have achieved convincing results. According to the decoding strategy, existing methods can be roughly divided into three categories: Connectionist Temporal Classiﬁcation (CTC) based~\cite{Shi2016An,Pan2016Reading,su2017accurate,wang2017gated}, attention mechanism based~\cite{shi2018aster,li2019show,baek2019wrong,wang2020decouple,yu2020towards,maurits2020bidirectional} and segmentation based~\cite{liao2019scene,wan2020textscanner}. From another perspective, recent methods can also be classified into autoregressive and non-autoregressive methods. Specifically, autoregressive methods decode the text from left to right, and the number of regressions depends on the length of the text. Most attention mechanism based methods adopt a left-to-right autoregressive decoding process. Although the performance is satisfactory, the inference speed is relatively slow, especially when dealing with long texts. On the contrary, non-autoregressive methods predict the text in parallel at a single time, the CTC-based methods are typical exemplars. Such parallel inference improves the speed significantly but ignores the dependencies between characters. We argue that fully non-autoregressive methods lack the context information of characters, and the assumption of independence increases the training difficulties of the hidden layers. Compared with autoregressive methods, the performance of non-autoregressive methods is relatively poor.

In this paper, we propose a parallel and iterative decoding model. In each iteration, the decoder still generates the text in parallel, and the context information is extracted depending on the previous predictions. Specifically, we adopt the easy first~\cite{2010yoavAn} decoding strategy according to the iterative generative process. Easy first predicts the most confident and obvious characters in each iteration first, and re-predicts other remaining characters in the next iteration. Different from traditional left-to-right decoding, easy first breaks the limitation of decoding order with higher flexibility. Our proposed parallel and iterative decoding strategy takes advantages of both fully autoregressive and fully non-autoregressive methods. Inspired by Transformer~\cite{2017vaswaniAttention}, the parallel decoder consists of a masked self-attention module, a 2D cross-attention module and a feed-forward network (FFN). 

\begin{figure*}[t]
\begin{center}
\includegraphics[width=0.9\linewidth]{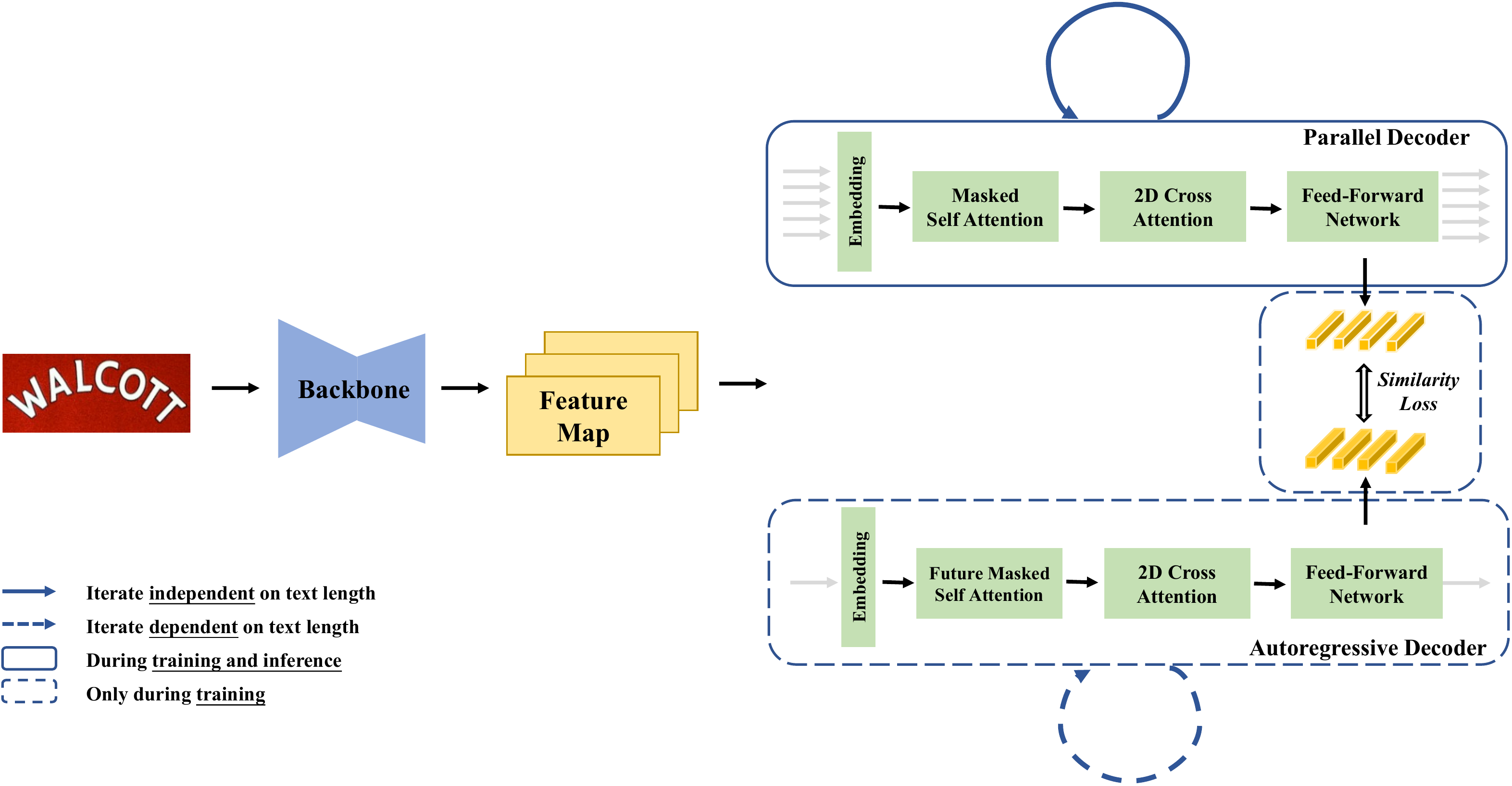}
\end{center}
   \caption{The framework of our proposed PIMNet, which contains three major components: an FPN based backbone, a parallel decoder and an autoregressive decoder. The parallel decoder adopts an iterative generation to extract the context information from the previous predictions. The autoregressive decoder shares the backbone with the parallel decoder and works as a teacher model to provide additional supervision signals to the hidden layer of parallel decoder. Note that the autoregressive decoder will be removed during inference for a faster speed.}
   \label{fig_pipeline}
\end{figure*}

Furthermore, although the iterative prediction brings the context information into the decoding, we observe that the prediction tends to be fully parallel in the early iterations. In addition to the weak supervision of the recognition labels, the assumption of independence may further increase the learning difficulties of hidden layers.
To address this problem, we propose mimicking learning into the existing framework. The mimicking learning brings additional supervision signals into the hidden layer (i.e., FFN.), which benefit learning of the hidden layer, especially reduce the learning difficulties caused by the strong assumption. Different from the hard recognition labels, the output of FFN contains more knowledge about visual and linguistic information, which can supervise the parallel decoding among the feature level. To implement the mimicking learning, we adopt an autoregressive decoder into the framework as a teacher model, where the autoregressive decoder shares the backbone with the parallel decoder. In this way, the whole model can be trained end-to-end without any pre-training of the teacher model, which significantly improves the training efficiency. During training, the output of FFN in the parallel decoder will fit the corresponding output of the autoregressive decoder. Based on the experiments, the mimicking makes the FFN outputs more distinguishable and further improves the recognition performance. Note that the autoregressive branch will be removed during inference to achieve a fast speed. In this paper, we call our proposed method \textbf{P}arallel, \textbf{I}terative and \textbf{M}imicking Network (PIMNet), and the main framework of PIMNet is shown in Fig.~\ref{fig_pipeline}.

As shown in Fig.~\ref{fig_acc-time}, 1-iteration PIMNet achieves the fastest inference speed due to the fully non-autoregressive decoding, but unsatisfactory accuracies. With the iterations of 5, the accuracies are improved significantly, since it overcomes some essential drawbacks of parallel decoding. Compared with other autoregressive methods, PIMNet takes a much shorter inference time with fewer iterations and achieves better or comparable performance. In summary, our PIMNet balances the recognition accuracy and inference speed using a novel parallel and iterative decoding.

The main contributions of our work are as follows:

1) We propose a parallel and iterative decoding framework for text recognition. Different from previous works, our method decodes with constant iterations independent on the text length, which takes the advantages of both fully autoregressive and fully non-autoregressive methods, and achieves a good balance between accuracy and efficiency.

2) To achieve the parallel and iterative decoding, an easy first strategy is designed for text recognition. Different from traditional left-to-right decoding, the easy first strategy predicts the most confident characters in each iteration, which is more flexible.

3) We propose to use mimicking learning to further help the training of the parallel decoder, where an additional autoregressive decoder is adopted as a teacher. To the best of our knowledge, we are the first to explore the mimicking learning between non-autoregressive and autoregressive decoding for text recognition.

4) Extensive experiments are conducted to verify the effectiveness and efficiency of the proposed method, which achieves state-of-the-art or comparable accuracies on popular benchmarks with a faster inference speed.

\vspace{-5pt}
\section{Related Work}

Scene text recognition has been studied for many years, and existing methods can be divided into traditional and deep learning based methods. Traditional methods~\cite{wang2010word, wang2011end, wang2012end, mishra2012scene,mishra2012top,novikova2012large,yao2014strokelets,neumann2012real} usually adopt a bottom-up framework, which detects and classifies characters first, then groups them with the lexicon, language model or heuristic rules. In recent years, deep learning based methods have dominated this area due to the simple pipeline and convincing performance. Based on the predicting strategies, existing methods can be roughly divided into non-autoregressive and autoregressive methods as follows.

\subsection{Non-Autoregressive Text Recognition}

Non-autoregressive methods aim to generate the target text at a single iteration or constant time independent on the text length. \cite{Jaderberg2016Reading} regards word recognition as classification with CNN, which lacks scalability due to the fixed vocabulary. Recent methods prefer to regard text recognition as a sequence-to-sequence task, and there are three main technologies for non-autoregressive decoding: CTC based, segmentation based and parallel-attention based. CTC-based methods~\cite{Shi2016An,Pan2016Reading,wang2017gated,su2017accurate,hu2020gtc,feng2019textdragon,chao2020variational} adopt CNN to extract visual features and CTC to transcribe the final text with short inference time. Segmentation-based methods~\cite{liao2019scene, wan2020textscanner} treat the recognition as semantic segmentation of characters, and they need additional character-level annotations. As for parallel-attention based methods, \cite{yu2020towards} proposes a novel parallel visual attention module to localize all characters in parallel and a global semantic reasoning module to capture global semantic context. However, 
the context information is limited by the predictions of the parallel visual attention module, and the context information may tend to further accumulate errors due to the wrong predictions. Apart from these three main categories, \cite{xie2019aggregation} proposes the aggregation cross-entropy loss with a new perspective of counting.

\subsection{Autoregressive Text Recognition}

Autoregressive methods always adopt an encoder-decoder framework, which predicts the sequence from left to right. Most attention-based methods belong to the autoregressive model, and they can be divided into 1D attention based and 2D attention based. For 1D attention based, \cite{lee2016recursive} proposes a recursive CNN network to capture broader features and an attention-based decoder to transcribe sequence. \cite{cheng2017focusing} introduces the problem of attention drift, and proposes focusing attention network to solve it. \cite{fang2018attention} proposes a fully CNN-based network to extract visual and language features separately. \cite{shi2016robust, shi2018aster,luo2019moran} rectify irregular text first then recognize it with 1D attention-based decoder. \cite{zhan2019esir} and \cite{yang2019symmetry} improve the quality of rectification with iteration and additional geometrical constraints respectively. For 2D attention based, \cite{yang2017learning} first introduces 2D attention into irregular text recognition and proposes an auxiliary segmentation task. \cite{li2019show} proposes an tailored 2D attention operation. 

Apart from previous works, \cite{wang2020decouple} decouples the prediction of attention weights and the autoregressive decoding process. \cite{luo2020learn} proposes a joint learning method for flexible data augmentation. \cite{qiao2020seed} introduces semantic global information to guide the decoding. \cite{litman2020scatter} trains a deep BiLSTM encoder to extract broader contextual dependencies. To solve the problem of attention diffusion, \cite{qz2} proposes a Gaussian constrained refinement module to refine the attention distribution. \cite{zheng2020lal} adopts an external language model to introduce useful context information into the decoding. \cite{wang2020exploring} aims at font styles of text recognition, and explores font-independent features. \cite{zhang2020autostr} first proposes to search data-dependent backbones for text recognition with neural architecture search.To handle the degradation text images, \cite{mou2020plugnet,wang2020scene} introduce the concept of super-resolution into text recognition.

In summary, autoregressive models usually achieve better performance but the left-to-right generation limits the inference speed. On the contrary, non-autoregressive models have faster inference speed but the accuracy may be lower. Nowadays, non-autoregressive decoding has been widely concerned in neural machine translation~\cite{gu2017non,wang2019non,2019ghazininejadAn}, auto speech recognition~\cite{chan2020imputer,chi2021align,tian2020spike} and image caption~\cite{guo2020non}. In the next section, we will introduce our proposed method which tries to balance the fully autoregressive and fully non-autoregressive models.


\begin{figure}[t]
\begin{center}
\includegraphics[width=1.0\linewidth]{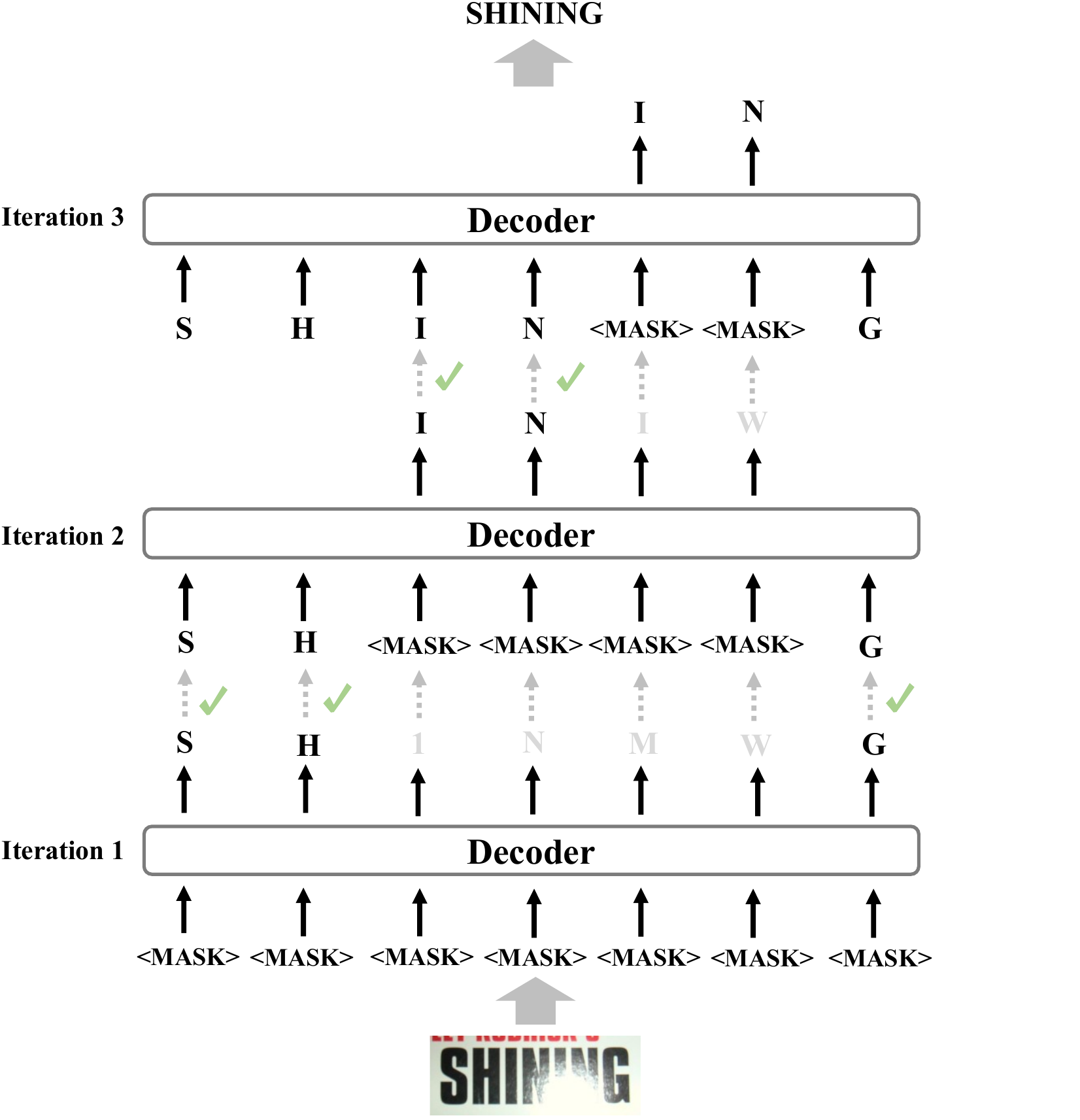}
\end{center}
   \caption{An illustration of easy first decoding strategy. In each iteration, the characters in black represent the characters with high confidence and will be reserved in the next iteration. The characters with low confidence will be replaced with the $\left\langle\texttt{MASK}\right\rangle$ token and re-predicted again based on the other reachable predictions.}
   \label{fig_easy_first}
\end{figure}

\section{Method}

In this section, we will describe our proposed PIMNet in detail. As shown in Fig.~\ref{fig_pipeline}, PIMNet consists of three major parts: FPN based Backbone, Parallel Decoder and Autoregressive Decoder. In Sec.~\ref{sec_3_1} we first introduce the backbone for feature extraction. The iterative decoding process and parallel decoder are introduced in Sec.~\ref{sec_3_2} and Sec.~\ref{sec_3_3} respectively. The autoregressive decoder and the details of mimicking learning will be described in Sec.~\ref{sec_3_4}. Finally, in Sec.~\ref{sec_3_5} we will introduce the objective function and the training strategy. 

\subsection{Backbone}
\label{sec_3_1}

Nowadays, some works~\cite{yang2019symmetry,wang2020decouple,yu2020towards} adopt FPN as a backbone for text recognition to extract multi-scale features. The fusion of the multi-scale features helps the model distinguish small and dense characters and be more robust to the background interference. Same as \cite{yang2019symmetry,yu2020towards}, we choose FPN equipped with ResNet-50~\cite{he2016deep} as our backbone. The output is the feature map from stage-3 with 1/8 size of the input image. Inspired from semantic segmentation~\cite{wang2018non-local,fu2019dual}, we stack two additional transformer units~\cite{2017vaswaniAttention} consisting of a self-attention module and a feed-forward network (FFN) to help the model extract broader features and distinguish the foreground and background. Finally, the enhanced feature maps are generated with a channel of 512 for the next two modules. 

\subsection{Iterative Decoding Strategy}
\label{sec_3_2}

\textbf{Easy First.} Easy first~\cite{2010yoavAn} is an iterative decoding strategy, where the most confident predictions in each iteration are first predicted. Inspired by~\cite{2019devlinBERT}, we adopt a special token $\left\langle\texttt{MASK}\right\rangle$ for the iterative decoding, which acts as a placeholder for the next iteration. As shown in Fig.~\ref{fig_easy_first}, in the first iteration all characters are unreachable, so the decoder is fed with $y^0_t=\left\langle\texttt{MASK}\right\rangle$ for all time steps $t$. 

The next iterations can be separated into two major steps of \textit{prediction} and \textit{update}:

1) \textit{Prediction}: In the current iteration, the parallel decoder predicts the corresponding probabilities of characters for each position which is still $\left\langle\texttt{MASK}\right\rangle$. On the contrary, the characters which have been already updated will not be re-predicted again:

\begin{equation}
\label{eq_easy_first_1}
    \hat{y}^i_t=
    \begin{cases}
    argmax(P(c_t|y^{i-1})),& y^{i-1}_t is \left\langle\texttt{MASK}\right\rangle\\
    y^{i-1}_t,& \textit{otherwise}
    \end{cases}
\end{equation}
where $\hat{y}^i_t$ indicates the character of time step $t$ in iteration $i$, $P(c_t)$ are the predicted probabilities for all characters categories at time step $t$, and $argmax(P)$ is the character with the largest probability. The conditional probability in $P$ indicates that the context information has been fully utilized based on the predictions from the previous iteration in a bi-directional manner.

2) \textit{Update}: After prediction, some positions of the target text are updated with the most confident predictions, and the others are abandoned and replaced by $\left\langle\texttt{MASK}\right\rangle$ again:

\begin{equation}
\label{eq_easy_first_2}
    y^i_t=
    \begin{cases}
    \hat{y}^i_t,& t\in top_k(max(P(c_m|y^{i-1})))\\
    \left\langle\texttt{MASK}\right\rangle,& \textit{otherwise}
    \end{cases}
\end{equation}
where $y^i$ is the final predictions in this iteration and will act as the input in the next iteration, $max(P)$ is the maximum probability for all candidate characters, $m$ indicates all the $\left\langle\texttt{MASK}\right\rangle$ positions in the $(i-1)th$ iteration.
$top_k$ represents the most confident $k$ predictions in the $m$ positions which are picked to update the previous predictions. $k$ is equal to $\lceil L/K \rceil$, where $L$ is the max length of the target text, and $K$ is the iteration number. 

In summary, the predicted text is updated with several of the most confident characters in each iteration, and the others are replaced by $\left\langle\texttt{MASK}\right\rangle$ which will be predicted in the next iterations. Note that, once a character is reserved and updated, it won't be updated again in the next iterations.

\begin{figure}[ht]
\begin{center}
\includegraphics[width=0.8\linewidth]{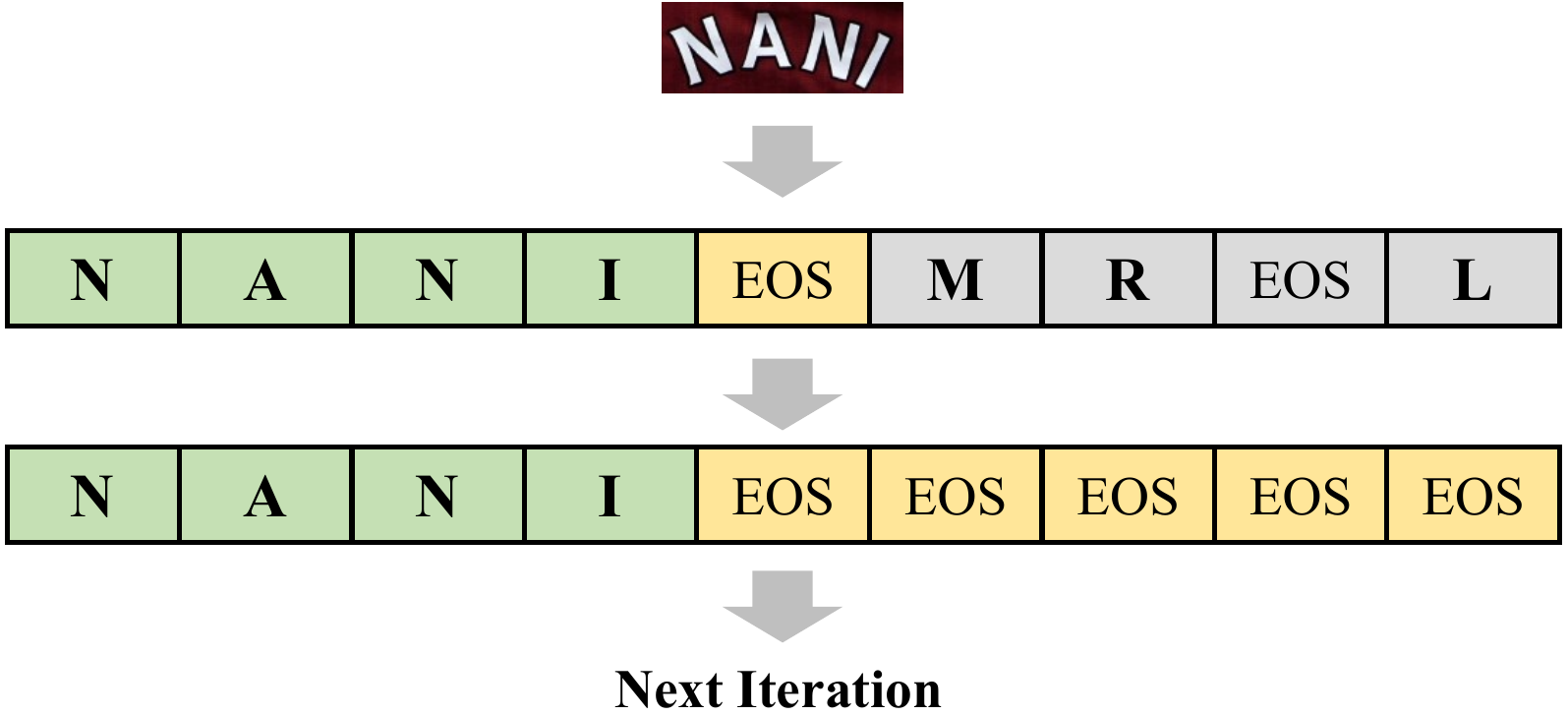}
\end{center}
   \caption{A simple illustration for determining the length of text.}
   \label{fig_eos_force}
\end{figure}

\noindent\textbf{About the Text Length.} In the implementation, the model needs to specify the maximum text length in advance. For autoregressive methods, an additional token $\left\langle\texttt{EOS}\right\rangle$ indicates the ending of the generated text, and the predicted characters after $\left\langle\texttt{EOS}\right\rangle$ are random and ignored. In traditional left-to-right decoding, the decoder only attends to the previous ones, and the random predicted characters cannot affect the prediction of the current time step. In parallel decoding, it is important to determine the valid length of the target text and prevent the decoder from attending to the random characters. In this paper, we propose an additional post-processing to deal with it, where we let the decoder predict $\left\langle\texttt{EOS}\right\rangle$ token and choose the first $\left\langle\texttt{EOS}\right\rangle$ token from left to determine the length of text. The rest characters are all replaced with $\left\langle\texttt{EOS}\right\rangle$ as shown in Fig.~\ref{fig_eos_force}. The predictions after the first $\left\langle\texttt{EOS}\right\rangle$ token will be ignored with mask operation, so the redundant random predictions will not affect the next parallel decoding in this way.

\subsection{Parallel Decoder}
\label{sec_3_3}

We adopt Transformer~\cite{2017vaswaniAttention}-based decoder as our parallel decoder without any RNN structure for high parallelism. It contains three main components: a masked self-attention, a 2D cross-attention, and an FFN. 

The input of masked self-attention is the embedding of the characters from the previous iteration combined with position encoding, which is mixed with the $\left\langle\texttt{MASK}\right\rangle$ tokens. Different from the original Transformer, our parallel decoder is bi-directional thus removes the original future-characters mask for the left-to-right generation. However, we add another mask to prevent the model from attending to the additional $\left\langle\texttt{MASK}\right\rangle$ tokens. The masked self-attention can extract abundant context information which is ignored in the fully non-autoregressive models. 

The 2D cross-attention is similar to the original Transformer, where the multi-head scale-dot attention operation is adopted. Specifically, the attention weights are calculated between the outputs of masked self-attention and the 2D feature map. The details of the calculation are the same as the Transformer, so we don't describe it in detail. Finally, the FFN introduces the non-linear transformation into the outputs of 2D cross-attention, and new predictions are generated later with a simple linear function in parallel. Same as the original Transformer, the residual connections exist between each module.

\subsection{Mimicking Learning for Parallel Decoder}
\label{sec_3_4}

In order to improve the learning of parallel decoder, we propose to adopt a mimicking learning process during training. Specifically, an additional autoregressive decoder is adopted as a teacher model, which shares the backbone with the parallel decoder. Similar to the parallel decoder, the teacher autoregressive decoder takes the Transformer-based structure consisting of a masked self-attention, a 2D cross attention and an FFN as shown in Fig.~\ref{fig_pipeline}. Note that the embedding of the characters fed into the masked self-attention is right-shifted, and the masked self-attention uses a future mask to make the model only attend to the previous characters, which introduces a left-to-right generating process for the decoder.

Different from some previous works about the  distillation~\cite{sanh2019distilbert,ydb1,jiao2020tinybert,sun2020mobilebert,ydb2}, the teacher decoder shares the backbone with the student decoder and the end-to-end training is achieved without any pre-training process of the teacher model, which improves the training efficiency. Besides the joint training with the shared feature map~\cite{hu2020gtc}, the mimicking learning provides more supervisions and transfers more knowledge to the parallel decoder, which improves the performance accordingly. Specifically, the FFN Mimicking is proposed, where the parallel decoder mimics the autoregressive decoder through the outputs of FFN. The details of FFN Mimicking will be described below.

FFN is the final hidden layer of the decoder, and it consists of abundant visual and linguistic features from the previous attention layers. The autoregressive decoder can extract more accurate and discriminative information because of the assumption of conditional dependence. Different from recognition labels, the outputs of FFN contain more knowledge and are essential to the final predictions. Mimicking the outputs of FFN can benefit the learning of the parallel decoder. Let $f^{par}$ be the outputs of the FFN in the parallel decoder and $f^{at}$ be the ones from the autoregressive decoder, the cosine-similarity based objective function is adopted as follows:

\begin{equation}
    L_{ffn} = 1-cosine(f^{par}, f^{at})\,.
\end{equation}

In the implementation, the gradients from the $L_{ffn}$ to the autoregressive decoder are stopped, i.e., the autoregressive decoder cannot be affected by the FFN Mimicking. In this way, the autoregressive decoder will achieve a more stable optimization.

\subsection{Objective Function and Training Strategies}
\label{sec_3_5}

Our proposed PIMNet is end-to-end trained with the objective function:

\begin{equation}
    L = \lambda_1L_{nat} + \lambda_2 L_{at}+\lambda_3L_{ffn}\,.
\end{equation}
$L_{nat}$ and $L_{at}$ are both standard cross-entropy losses of the predicted probabilities with respect to the label of transcription, where $L_{nat}$ is the loss of parallel decoder and $L_{at}$ is the loss of the autoregressive decoder. $\lambda$ is used to balance the losses, and we set $\lambda_1=\lambda_2=\lambda_3=1$ in our implementation.

During training, we adopt the teacher forcing strategy for both parallel decoder and autoregressive decoder. For the autoregressive decoder, it is same with previous works. For the parallel decoder, in each iteration, the updating strategy shown in Eq.~\ref{eq_easy_first_2} can be extended as follows:

\begin{equation}
\label{eq_teacher_force}
    y^i_{update,t}=
    \begin{cases}
    label_t,& \textit{training}\\
    \hat{y}^i_t,& \textit{inference}
    \end{cases}
\end{equation}
where $y^i_{update,t}$ is the position with high confidence and ready to be updated. We use the corresponding characters from the ground truth instead of the prediction to update the target text in each iteration, which leads to faster and better convergence.


\begin{table*}[ht]
    \caption{Lexicon-free performance on public benchmarks. \textbf{Bold} represents the best performance. \underline{Underline} represents the second best performance. * represents the performance of the position enhancement branch, which is a non-autoregressive decoder. \dag denotes the performance is from the same number of attention operations as ours for a fair comparison. The inference time is measured under the same software and hardware environments with the official implementations. ``Real" is the public real training data.}
    \label{tabel_sota}
    \centering
    \small
    \begin{tabular}{cl|c|c|c|cc|cc|c|c|c|c}
    \hline 
    & \multirow{2}{*}{Methods} & \multirow{2}{*}{Training Data} & IIIT5K & SVT & \multicolumn{2}{c|}{IC13} & \multicolumn{2}{c|}{IC15} & SVTP & CUTE & \multirow{2}{*}{Average} & Time \cr
    &                          & &3000   & 647 & 857 & 1015               & 1811 & 2077              & 645  & 288  & & ms/image \cr
    \hline
    \hline
    \parbox[t]{2mm}{\multirow{11}{*}{\rotatebox[origin=c]{90}{\textbf{Autoregressive}}}}
    & Shi \textit{et al.}~\cite{shi2018aster}        & 90K + ST & 93.4 & 89.5 &  -   & 91.8 & 76.1 &   -  & 78.5 & 79.5 & 86.8 & 77.3\\
    & Zhan \textit{et al.}~\cite{zhan2019esir}       & 90K + ST & 93.3 & 90.2 &  -   & 91.3 & 76.9 &   -  & 79.6 & 83.3 & 87.2 & -\\
    & Li \textit{et al.}~\cite{li2019show}           & 90K + ST & 91.5 & 84.5 &  -   & 91.0 &   -  & 69.2 & 76.4 & 83.3 & 83.2 & 45.4\\
    & Luo \textit{et al.}~\cite{luo2019moran}        & 90K + ST & 91.2 & 88.3 &  -   & 92.4 & 74.7 & 68.8 & 76.1 & 77.4 & 81.6 & 57.9\\
    & Yang \textit{et al.}~\cite{yang2019symmetry}   & 90K + ST & 94.4 & 88.9 &  -   & 93.9 & 78.7 &  -   & 80.8 & \underline{87.5} & 88.5 &- \\
    & Wang \textit{et al.}~\cite{wang2020decouple}   & 90K + ST & 94.3 & 89.2 &  -   & 93.9 &   -  & 74.5 & 80.0 & 84.4 & 86.9 &-\\
    & Qiao \textit{et al.}~\cite{qiao2020seed}       & 90K + ST & 93.8 & 89.6 &  -   & 92.8 & \underline{80.0} &  -   & 81.2 & 83.6 & 88.4 & 80.1\\
    & Wang \textit{et al.}~\cite{wang2020exploring} & 90K + ST & 94.4 & 89.8 & - & 93.7 & - & 75.1 & 80.2 & 86.7 & 87.2 & -\\
    & Yue \textit{et al.}~\cite{yue2020robust}       & 90K + ST & \textbf{95.3} & 88.1 &  -   & \underline{94.8} &   -  & 77.1 & 79.5 & \textbf{90.3} & 88.2 & -\\
    & Zhang \textit{et al.}~\cite{zhang2021spin} & 90K + ST & \underline{95.2} & \underline{90.9} & - & \underline{94.8} & \textbf{82.8} & \underline{79.5} & \underline{83.2} & \underline{87.5} & \textbf{90.4} & -\\
    & Mou \textit{et al.}~\cite{mou2020plugnet}      & 90K + ST & 94.4 & \textbf{92.3} &  -   & \textbf{95.0} &  -   & \textbf{82.2} & \textbf{84.3} & 85.0 & \underline{89.8} & 61.3\\
    \hline
    \parbox[t]{2mm}{\multirow{8}{*}{\rotatebox[origin=c]{90}{\textbf{Non-Autoregressive}}}}
    & Liao \textit{et al.}~\cite{liao2019scene}      &  ST & 91.9 & 86.4 &   -  & 91.5 &  -   &  -   &  -   & 79.9 &  \underline{90.4} &-\\
    & Yu \textit{et al.}~\cite{yu2020towards}\dag        & 90K + ST & 93.8 & 88.5 & \underline{94.7} &  -   & 81.7 &  -   & 82.6 & \textbf{88.9} & 89.2 & 32.8\\
    & Yue \textit{et al.}~\cite{yue2020robust}*       & 90K + ST & 92.5 & 84.4 &  -   & 91.4 &   -  & 70.7 & 74.4 & 83.3 & 85.7 & -\\
    & PIMNet w/o Mimicking (Ours)                     & 90K + ST & \underline{94.9}  & \underline{90.1} & \underline{94.7}  & \underline{92.3} & \underline{82.9} & \underline{79.9}  & \underline{83.4} & \underline{86.1} & 90.3 & 28.4\\
    & PIMNet                (Ours)                   & 90K + ST & \textbf{95.2}  & \textbf{91.2} & \textbf{95.2}  & \textbf{93.4} & \textbf{83.5} & \textbf{81.0}  & \textbf{84.3} & 84.4 & \textbf{90.9} & 28.4\\
    & Hu \textit{et al.}~\cite{hu2020gtc} & 90K + ST + Real & 95.8 & 92.9 & - & 94.4 & - & 79.5 & 85.7 & 92.2 & 90.0 & 21.6\\
    & Wan \textit{et al.}~\cite{wan2020textscanner} & 90K + ST + Real& 95.7 & 92.7 &   -  & 94.9 & 83.5 &  -   & 84.8 & 91.6 & \underline{91.2} &  -\\
    & PIMNet                (Ours)                   & 90K + ST + Real & \textbf{96.7}  & \textbf{94.7} & \textbf{96.6}  & \textbf{95.4} & \textbf{88.7} & \textbf{85.9}  & \textbf{88.2} & \textbf{92.7} & \textbf{93.8} & 28.4\\
    \hline
    \end{tabular}
\end{table*}

\section{Experiment}

\subsection{Datasets}

\noindent\textbf{IIIT5K-Words (IIIT5K)} \cite{mishra2012scene} contains 5000 images collected from website. There are 3000 images for testing, most of which are horizontal with high quality.

\noindent\textbf{Street View Text (SVT)} \cite{wang2011end} consists of 647 cropped word images from 249 street view images, which targets at regular text recognition.

\noindent\textbf{SVT-Perspective (SVTP)} \cite{quy2013recognizing} comprises 645 word images cropped from SVT, which is usually used for evaluating the performance of recognizing perspective text.

\noindent\textbf{ICDAR2013 (IC13)} \cite{karatzas2013icdar} consists of 1095 regular-text images for testing. Two different versions are used for evaluation: 1015 and 857 images, which discard images that contain non-alphanumeric characters and less than three characters respectively.

\noindent\textbf{ICDAR2015 (IC15)} \cite{karatzas2015icdar} is a challenging dataset for recognition due to the degraded images collected without careful focusing. It contains 2077 cropped images, and some works used 1811 images for evaluation without some extremely distorted images.

\noindent\textbf{CUTE80 (CUTE)} \cite{risnumawan2014robust} consists of 288 word images, which is used for irregular text recognition. Most of them are curved, and no lexicon is provided.

\noindent\textbf{Synth90K (90K)} \cite{Jaderberg2016Reading} is a synthetic dataset widely used for training recognition models, which contains 9 million synthetic images.

\noindent\textbf{SynthText (ST)} \cite{gupta2016synthetic} is another synthetic dataset for text detection task, which contains some distorted or curved text instances. We cropped them with the bounding boxes provided by ground-truth.

\subsection{Implementation Details}

\noindent \textbf{Model Settings}: There are 37 symbols covered for recognition, including numbers, lower case characters and $\left\langle\texttt{EOS}\right\rangle$ token. The details of the transformer units are illustrated in Tab.~\ref{tabel_params}, where $N$ is the number of Transformer units, $d_{ffn}$ is the dimension of the hidden layer in FFN, $h$ is the number of attention heads. The number of iterations of the parallel decoder is set to 5.

\begin{table}[h]
\caption{The hyper-parameters of the Transformer units adopted in different modules.}
\label{tabel_params}
\centering
   \begin{tabular}{l|ccc}
   \hline 
    & $N$ & $h$ & $d_{ffn}$ \\
   \hline
   Backbone & 2 & 8 & 512\\
   Parallel Decoder & 1 & 8 & 1024\\
   Autoregressive Decoder & 1 & 8 & 1024 \\
   \hline  
   \end{tabular}
\end{table}

\noindent \textbf{Training}: The model is trained with two synthetic datasets 90K and ST from scratch, the input images are resized to $64\times256$ directly. To compare with some methods fairly, we also train our model with additional real training data provided from public benchmarks (IIIT5K, SVT, IC03, IC13, IC15, COCO). Adam~\cite{kingma2014adam} is adopted as our optimizer with the initial learning rate $10^{-4}$. The batch size is set to 256, and the model is trained for nearly 8 epochs. We adopt rotation, perspective distortion, motion blur, and Gaussian noise for data augmentation. 

\noindent \textbf{Inference}: We directly use the model trained on two synthetic datasets for evaluation without fine-tuning. The input images are still resized to $64\times256$, and the beam search is not adopted. 

\noindent \textbf{Implementation}: Our model is implemented with TensorFlow~\cite{2016tensorflow}. The training and evaluation are both conducted on a single NVIDIA M40 graphics card.

\begin{table*}[t]
\caption{The comparison of accuracy and inference time with different iterations.}
\label{tabel_iteration}
\centering
\small
   \begin{tabular}{l|c|c|c|c|c|c|c}
   \hline 
   Iterations &  IIIT5K & SVT  & IC13 & IC15 & SVTP & CUTE & Time (ms/image)\\
   \hline
   1          &  93.1            & 88.7          & 90.8         & 76.3         & 80.0          & 82.3 & 17.6\\
   2          &  94.3            & 90.0          & 92.6         & 78.8         & 81.7          & 85.1 & 20.9\\
   3          &  94.6            & 90.1          & 92.8         & 79.3         & 82.6          & \textbf{86.8} & 24.7\\
   5          &  94.8            & 90.4          & 93.2         & 79.7         & 83.3          & 86.5 & 28.4\\
   6          &  94.8            & 90.4          & 93.6         & 79.9         & 83.2          & 86.2 & 30.7\\
   10         &  95.0            & 90.3          & \textbf{93.7}& 80.5         & 85.1          & 85.8 & 42.6\\
   30         &  \textbf{95.8}   & \textbf{90.7} & 93.5         & \textbf{80.6}& \textbf{85.7} & 86.5 & 87.9\\
   \hline  
   \end{tabular}
\end{table*}

\subsection{Comparisons with State-of-the-Arts}

As shown in Tab.~\ref{tabel_sota}, we compare our PIMNet with other published state-of-the-art methods divided into autoregressive and non-autoregressive models. Compared with other non-autoregressive methods, our method achieves the 7 best results among 8 benchmarks. Specifically, PIMNet gets $1.4\%$, $2.7\%$ and $1.9\%$ improvements on IIIT5K, SVT and IC13 respectively, which shows the effectiveness of our method on regular text benchmarks. $1.8\%$ and $1.7\%$ improvements are achieved on two irregular datasets IC15 and SVTP without any pre-processing like rectification. Compared with the CTC-based method GTC~\cite{hu2020gtc}, our PIMNet outperforms it on all six benchmarks under the same setting of training data. Besides autoregressive guidance, our PIMNet also adopts an iterative easy first decoding strategy to extract context information and mimicking learning to improve the learning of the hidden layers, which is a further step. Compared with parallel-attention based method SRN~\cite{yu2020towards}, our PIMNet outperforms it on five benchmarks with the same number of attention operations, especially $2.7\%$ and $1.8\%$ improvements on SVT and IC15. Our proposed iterative generation can further improve the quality of context information with the most confident predictions in each iteration.

This paper tries to balance the accuracy and the efficiency, our PIMNet achieves comparable accuracy with most autoregressive methods at a much faster inference speed. Compared with the rectification-based method ASTER~\cite{shi2018aster}, our method runs nearly three times faster and gets better accuracy on all six benchmarks, especially $7.4\%$, $5.8\%$ and $4.9\%$ improvements on IC15, SVTP and CUTE respectively. PlugNet \cite{mou2020plugnet} is another state-of-the-art method with novel super-resolution units, and our method achieves more than two times faster inference speed and gets comparable performance and even better results on IIIT5K. 

\vspace{-3pt}
\subsection{Ablation Studies}
In this section, we conduct extensive ablation studies to further analyze our PIMNet from quantitative and qualitative perspectives.

\noindent\textbf{Analysis of Iteration Number.} The number of iteration is an important hyper-parameter in our method, so we conduct experiments to analyze the effect caused by the iterations number, and we remove the autoregressive decoder during training. As shown in Tab.~\ref{tabel_iteration}, the fully parallel model with only one iteration works poorly on six benchmarks, as we discussed before, the fully parallel decoding lacks the useful context information and impacts on the optimization. When an additional iteration is adopted, the performance is improved significantly, especially on IC13 (from $90.8\%$ to $92.6\%$), IC15 (from $76.3\%$ to $78.8\%$) and CUTE (from $82.3\%$ to $85.1\%$). With the increment of iterations, the accuracy and inference time increase accordingly. In order to achieve a balance, the number of iterations 5 is selected in the implementation. Note that the maximum text length in our implementation is set to 30, so the model with 30 iterations in the Tab.~\ref{tabel_iteration} generates the text like autoregressive decoding, which also shows the flexibility of our model between autoregressive and non-autoregressive models.

\begin{figure}[h]
\begin{center}

\includegraphics[width=1.0\linewidth]{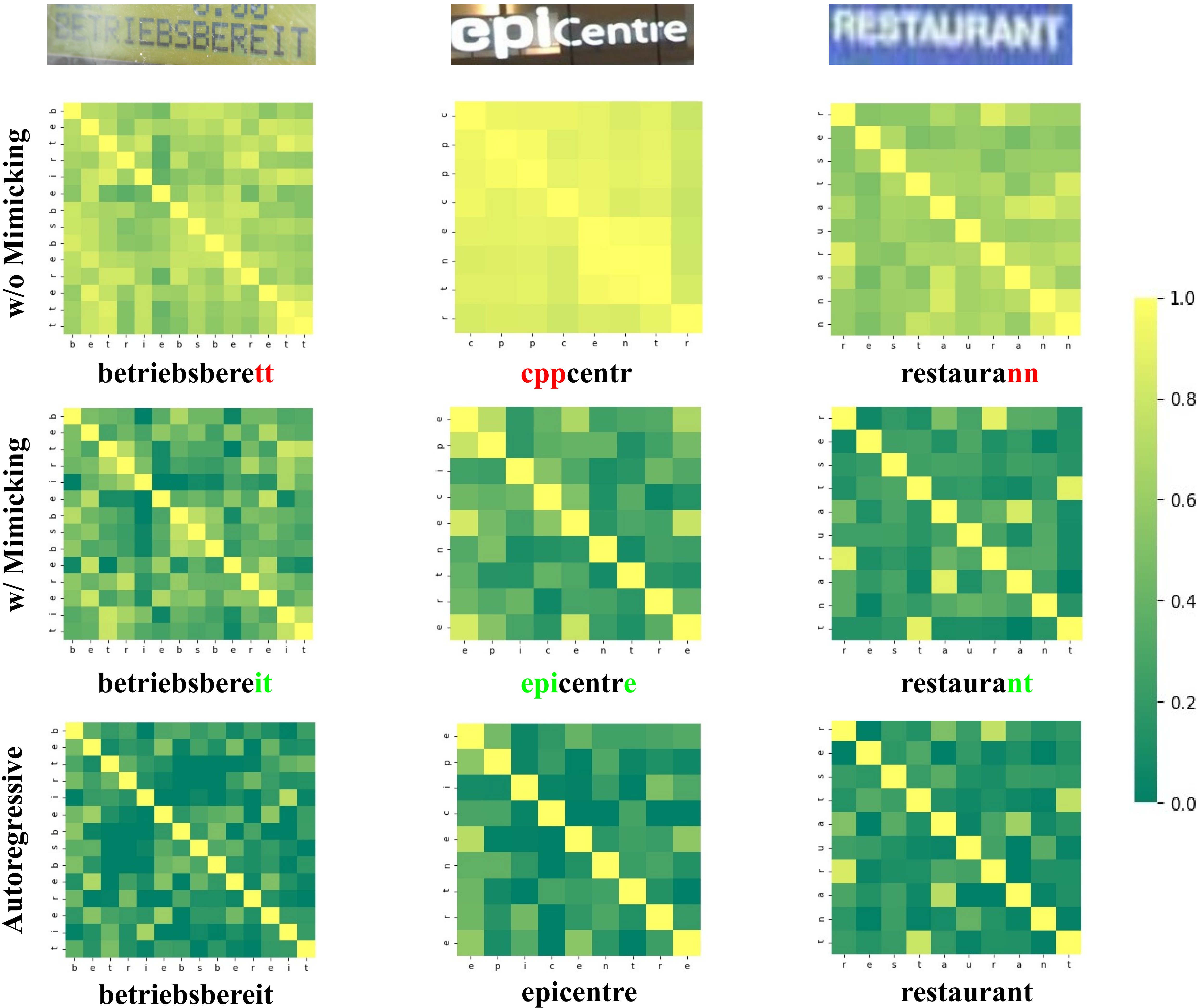}
\end{center}
   \caption{The normalized cosine similarities of FFN output. Each pixel shows the cosine similarities $cos_{ij}$ of the i-th and j-th outputs. With mimicking learning, the output features are more discriminative, and the duplicate or missing predictions of the parallel decoder can be prevented.}
   \vspace{-4pt}
   \label{fig_ffn_sim_vis}
\end{figure}

\noindent\textbf{Analysis of Mimicking Learning.} Mimicking learning is adopted to improve the learning of hidden layers in the parallel decoder. As shown in Tab.~\ref{tabel_sota}, PIMNet with mimicking learning achieves better accuracy, especially $1.1\%$ on SVT and $1.1\%$ on IC15. Note that the PIMNet without mimicking still retains the autoregressive decoder, which is similar to GTC~\cite{hu2020gtc}.

To further analyze the effectiveness of the mimicking learning, we visualize the cosine similarities of FFN outputs as shown in Fig.~\ref{fig_ffn_sim_vis}. We explain that the parallel decoder in the early iterations tends to predict similar outputs due to the fully parallel decoding, thus the similarities are large across different positions. The similar outputs of FFN may mislead the final predictions, which tend to be duplicate and wrong. To address this problem, mimicking learning introduces some dependencies into the parallel decoder from the autoregressive decoder. With more distinguishable FFN outputs, the predictions are improved. Compared with the autoregressive decoder, the cosine similarity matrices of PIMNet with mimicking are similar, which verifies the effectiveness of the mimicking learning.

\begin{table}[h]
\caption{The performance of PIMNet with or without the mimicking (``mimic'' for short) on different text lengths. ``Gap'' is the accuracy gain when using mimicking learning.}
\label{tabel_mimic_length}
\centering
\small
   \begin{tabular}{c|cccccccc}
   \hline
   \multirow{2}{*}{Methods} &  \multicolumn{8}{c}{Length}\cr
                            & $\leq3$ & 4 & 5 & 6 & 7 & 8 & 9 & $\geq10$ \cr
   \hline
   w/o mimic & \textbf{96.0} & 97.9 & 95.2 & \textbf{95.3} & 93.2 & 90.9 & 96.1 & 83.3 \\
   w/ mimic  & 95.4 & 97.9 & \textbf{95.8} & 94.7 & \textbf{94.5} & \textbf{94.6} & \textbf{96.9} & \textbf{85.3}\\
 
   \hline
   Gap & -0.6 & 0.0 & 0.6& -0.6& 1.3& 3.7& 0.8& 2.0 \\
   \hline
   \end{tabular}
\end{table}

In Tab.~\ref{tabel_mimic_length}, we compare the accuracies on different text lengths with or without mimicking learning. When dealing with the texts of length shorter than 6, the accuracy of both is similar with a gap less than 1. However, with the increment of length, the gap is also increasing, and the PIMNet with mimicking works better on long texts. One explanation is that the parallel decoder without mimicking tends to be confused on adjacent characters in long text due to the similar FFN outputs.

\begin{table}[h]
\caption{The comparison of inference time among different decoding strategies.}
\label{tabel_speed}
\centering
   \begin{tabular}{l|c}
   \hline 
    Methods & Time (ms/image) \\
   \hline 
   CTC based          &   16.3  \\
   1D Attention based &   50.4   \\
   2D Attention based &   57.5  \\
   \hline
   PIMNet (1 iteration) &   17.6 \\
   PIMNet (5 iterations)&   28.4  \\
 
   \hline  
   \end{tabular}
\end{table}

\noindent\textbf{Analysis of Inference Speed.} As shown in the Tab.~\ref{tabel_sota}, our proposed method achieves a faster inference speed than most recent models. In order to further verify our advantage in efficiency, we compare our method with CTC, 1D attention based and 2D attention based decoders. In order to remove the influence of the backbone, we evaluate all methods with the same backbone as our PIMNet. As shown in Tab.~\ref{tabel_speed}, our method with 5 iterations is almost 2 times faster than 1D attention and 2D attention based decoders. When the iteration number is 1, the inference time is close to CTC based decoder, since they are both fully parallel decoding. In other words, our method is flexible between efficiency and accuracy, so the iteration number can be modified simply according to different requirements in real applications.

\begin{table}[h]
\caption{The performance of PIMNet with or without the post-processing (PP) of text length.}
\label{tabel_eos}
\centering
\small
   \begin{tabular}{l|c|c|c|c|c|c}
   \hline
   Method &  IIIT5K & SVT  & IC13 & IC15 & SVTP & CUTE \\
   \hline
   w/o PP  &   94.2  & 89.2 & 92.4 & 77.6 & 82.5 & 82.6\\
   w/ PP &   \textbf{95.2}  & \textbf{91.2} & \textbf{93.4} & \textbf{81.0} & \textbf{84.3} & \textbf{84.4} \\
 
   \hline  
   \end{tabular}
\end{table}

\noindent\textbf{Analysis of Text Length Post-Processing.} In Sec.~\ref{sec_3_2}, we describe post-processing of text length to prevent the model from attending to the redundant random predictions. Here we compare the model with and without the post-processing to analyze the effectiveness. As shown in Tab.~\ref{tabel_eos}, the model with post-processing achieves much better performance on all six benchmarks. As we mentioned before, the redundant predicted characters will introduce some noises into the extraction of context information.


\begin{figure}[ht]
\begin{center}
\includegraphics[width=1.0\linewidth]{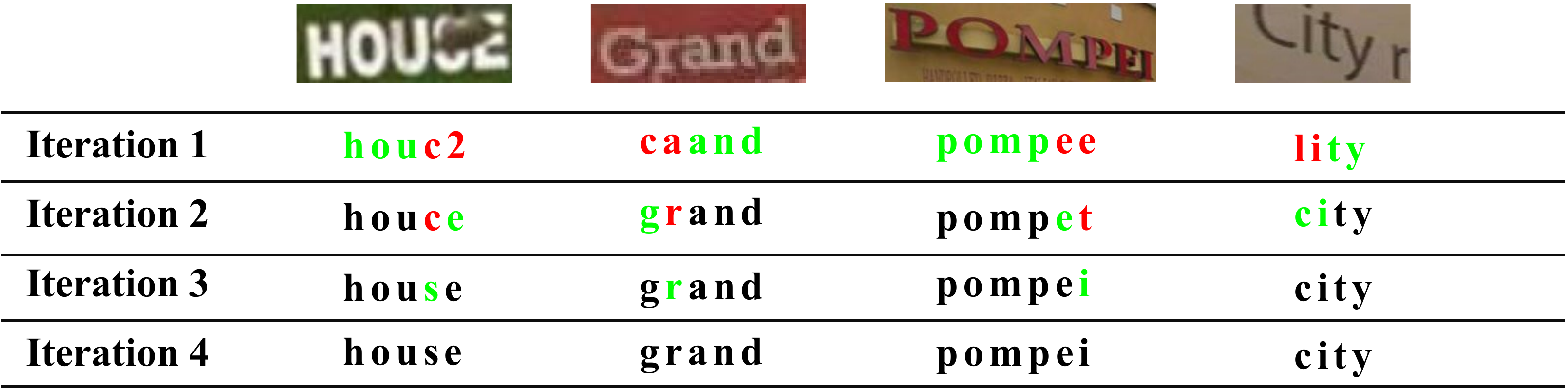}
\end{center}
   \caption{Some examples to illustrate the iterative generation with easy first. Characters in green denote the characters with high confidence and are used to update the target text. Oppositely, characters in red indicate the relatively low-confidence predictions, which are temporarily abandoned and re-predicted in the next iteration.}
   \label{fig_iteration_examples}
\end{figure}

\section{Qualitative Analysis}

As shown in Fig.~\ref{fig_iteration_examples}, we illustrate the process of iterative easy first decoding with some examples. In each iteration, the characters in green represent the most confident predictions, which are reserved to update the final predicted text. The characters in red are the predictions with relatively low confidence, which are abandoned in this iteration again. After each iteration, the context information within reachable characters is extracted and helps the model predict other characters in the next iteration. Easy first decoding breaks the restrictions of the left-to-right or the right-to-left reading orders, and it can predict some hard-distinguished characters in the later iterations flexibly. For example, in the first image, the occluded character ``s'' can be predicted after ``e'', which can bring more context information and benefit the recognition of ``s''. In the last image, the uncompleted character ``c'' can be refined with the help of the remaining characters.


\section{Conclusion}

In this paper, we propose a Parallel, Iterative and Mimicking Network (PIMNet) to balance the accuracy and efficiency of scene text recognition. PIMNet adopts a parallel decoder with an iterative generation. In each iteration, the easy first strategy is used, where several of the most confident characters are updated. With the iterative decoding process, the context information is fully explored. To improve the learning of the parallel decoder, we propose to use mimicking learning into the framework, where an additional autoregressive decoder is adopted as a teacher model. Furthermore, the parallel decoder mimics the output of the hidden layer of the autoregressive decoder. More supervision signals benefit the training of the parallel decoder, and the performance is further improved. PIMNet achieves state-of-the-art results on most benchmarks and a faster inference speed compared with other autoregressive methods. In summary, our method takes advantages of both the fully autoregressive and fully non-autoregressive models with more flexibility. In the future, we will extend our PIMNet with fewer parameters and propose a much lighter framework. Furthermore, we'd like to introduce a semi-~\cite{bartz2017see,tarvainen2017mean,rasmus2015semi}, weakly-~\cite{qxg2,zhong2020boosting,hsu2020learning} or self-supervised~\cite{ldz1,ldz2,ldz3,lxn1,lw1,zyc1,zyf1,zyf2} learning framework to improve the model with fewer labels.

\begin{acks}
Supported by the Open Research Project of the State Key Laboratory of Media Convergence and Communication, Communication University of China, China (No. SKLMCC2020KF004), the Beijing Municipal Science \& Technology Commission (Z191100007119002), the Key Research Program of Frontier Sciences, CAS, Grant NO ZDBS-LY-7024, the National Natural Science Foundation of China (No. 62006221), and CAAI-Huawei MindSpore Open Fund.
\end{acks}

\bibliographystyle{ACM-Reference-Format}
\balance
\bibliography{refer}










\end{document}